\documentclass[runningheads]{llncs}

\usepackage{times}  
\usepackage{helvet}  
\usepackage{courier}  
\usepackage{url}  
\usepackage{graphicx}  

\usepackage[utf8]{inputenc} 
\usepackage[T1]{fontenc}    
\usepackage{booktabs}       
\usepackage{amsfonts}       
\usepackage{microtype}      
\usepackage{latexsym}
\usepackage{algorithm}
\usepackage{algorithmic}
\usepackage{amsmath}
\usepackage[normalem]{ulem}
\usepackage{color}
\usepackage{caption,booktabs,array}
\usepackage{multirow}
\usepackage{subfig}
\usepackage[font=small]{caption}

\makeatletter
    \let\@internalcite\cite
    \def\cite{\def\citeauthoryear##1##2{##1, ##2}\@internalcite}
    \def\shortcite{\def\citeauthoryear##1{##2}\@internalcite}
    \def\@biblabel#1{\def\citeauthoryear##1##2{##1, ##2}[#1]\hfill}
\makeatother

\begin{document}

\title{Parametric t-Distributed Stochastic Exemplar-centered Embedding}

\author{Martin Renqiang Min\inst{1} \and
Hongyu Guo\inst{2} \and
Dinghan Shen\inst{3}}
\authorrunning{Min et al.}
\institute{NEC Labs America, Princeton, NJ 08540 \\
\email{renqiang@nec-labs.com} \and
National Research Council Canada, Ottawa, ON K1A 0R6 \\
\email{hongyu.guo@nrc-cnrc.gc.ca} \and
Duke University Durham, NC 27708\\
\email{dinghan.shen@duke.edu} 
}
\maketitle              

\begin{abstract}
Parametric embedding methods such as parametric t-distributed Stochastic Neighbor Embedding (pt-SNE) enables out-of-sample data visualization without further computationally expensive optimization or approximation. However, pt-SNE favors small mini-batches to train a deep neural network but large mini-batches to approximate its cost function involving all pairwise data point comparisons, and thus has difficulty in finding a balance. To resolve the conflicts, we present parametric t-distributed stochastic exemplar-centered embedding. Our strategy learns embedding parameters by comparing training data only with precomputed exemplars to indirectly preserve local neighborhoods, resulting in a cost function with significantly reduced computational and memory complexity. Moreover, we propose a shallow embedding network with high-order feature interactions for data visualization, which is much easier to tune but produces comparable performance in contrast to a deep feedforward neural network employed by pt-SNE. We empirically demonstrate, using several benchmark datasets, that our proposed method significantly outperforms pt-SNE in terms of robustness, visual effects, and quantitative evaluations.  
\end{abstract}
\section{Introduction}
Unsupervised nonlinear dimensionality reduction methods, which embed high-dimensional data to a low-dimensional space, have been extensively deployed in many real-world applications for data visualization. Data visualization is an important component of data exploration and data analytics, as it helps data analysts to develop intuitions and gain deeper understanding about the mechanisms underlying data generation. Comprehensive surveys about dimensionality reduction and data visualization methods can be found in van der Maaten et al. (2009)~\cite{Maaten08dimensionalityreduction} and Burges (2010)~\cite{dimension-reduction-a-guided-tour-2}. Among these approaches, nonparametric neighbor embedding methods such as t-SNE~\cite{van2008visualizing} and  Elastic Embedding~\cite{carreira2010elastic} are widely adopted. They generate low-dimensional latent representations by preserving neighboring probabilities of high-dimensional data in a low-dimensional space, which involves pairwise data point comparisons and thus has quadratic computational complexity with respect to the size of a given data set. This prevents them from scaling to any dataset with a size beyond several thousand. Moreover, these methods are not designed for readily generating the embedding of out-of-sample data that are prevalent in modern big data analytics. To generate out-of-sample data embedding given an existing sample embedding, computationally expensive numerical optimization or Nystr{\" o}m approximation is often performed, which is undesirable in practice~\cite{bengio2004out,vladymyrov2014linear,carreira2015fast}. 

Parametric embedding methods, such as parametric t-SNE (pt-SNE)~\cite{Maaten09} employing a deep neural network (DNN), learn an explicit parametric mapping function from a high-dimensional data space to a low-dimensional embedding space, which can readily generate the embedding of out-of-sample data. The objective function of pt-SNE is the same as that of t-SNE with quadratic computational complexity. Fortunately, owing to the explicit mapping function defined by the DNN, optimization methods such as stochastic gradient descent or conjugate gradient descent based on mini-batches can be deployed when pt-SNE is applied to large-scale datasets. 

However, on one hand, the objective function of pt-SNE is a sum of a quadratic number of terms over pairwise data points, which requires mini-batches with fairly large batch sizes to achieve a reasonably good approximation to the original objective; On the other hand, optimizing the parameters of the DNN in pt-SNE also requires careful choices of batch sizes, which is often best served with small batch sizes to avoid being stuck in a bad local minimum. These conflicting choices of batch sizes make the optimization of pt-SNE hard and render its performance sensitive to the chosen batch size. In addition, to approximate the loss function defined over all pairwise data points, pt-SNE independently computes pairwise neighboring probabilities of high-dimensional data for each mini-batch, so it often produces dramatically different embeddings with different choices of user-defined perplexities that are coupled with batch sizes. Finally, although the mapping function of pt-SNE parameterized by a DNN is powerful, it is very hard to learn and requires complicated procedures such as tuning network architectures and tuning many hyper-parameters. For data embedding and visualization purposes, most users are reluctant to go through these complicated procedures.

To address the aforementioned problems, in this paper, we present unsupervised parametric t-distributed stochastic exemplar-centered embedding. Instead of modeling pairwise neighboring probabilities, our strategy learns embedding parameters by comparing high-dimensional data only with precomputed representative high-dimensional exemplars, resulting in an objective function with linear computational and memory complexity with respect to the number of exemplars. The exemplars are identified by a small number of iterations of k-means updates, taking into account both local data density distributions and global clustering patterns of high-dimensional data. These nice properties make the parametric exemplar-centered embedding insensitive to batch size and scalable to large-scale datasets. All the exemplars are repeatedly included into each mini-batch, and the choice of the perplexity hyper-parameter only concerns the expected number of neighboring exemplars calculated globally, independent of batch sizes. Therefore, the perplexity is much easier to choose by the user and much more robust to produce good embedding performance. We further use noise contrastive samples to avoid comparing data points with all exemplars, which further reduces computational/memory complexity and increases scalability. Although comparing training data points only with representative exemplars indirectly preserves similarities between pairwise data points in each local neighborhood, it is much better than randomly sampling small mini-batches in pt-SNE whose coverages are too small to capture all pairwise similarities on a large dataset.

Moreover, we propose a shallow embedding network with high-order feature interactions for data visualization, which is much easier to tune but produces comparable performance in contrast to a deep neural network employed by pt-SNE. Experimental results on several benchmark datasets show that, our proposed parametric exemplar-centered embedding methods for data visualization significantly outperform pt-SNE in terms of robustness, visual effects, and quantitative evaluations. We call our proposed deep t-distributed stochastic exemplar-centered embedding method dt-SEE and high-order t-distributed exemplar-centered embedding method hot-SEE.

Our contributions in this paper are summarized as follows: (1) We propose a scalable unsupervised parametric data embedding strategy with an objective function of significantly reduced computational complexity, avoiding pairwise training data comparisons in existing methods; (2) With the help of exemplars, our methods eliminate the instability and sensitivity issues caused by batch sizes and perplexities haunting other unsupervised embedding approaches including pt-SNE; (3) Our proposed approach hot-SEE learns a simple shallow high-order parametric embedding function, beating state-of-the-art unsupervised deep parametric embedding method pt-SNE on several benchmark datasets in terms of both qualitative and quantitative evaluations.

\section{Related Work}\label{sec:related}
Dimensionality reduction and data visualization have been extensively studied in the last twenty years~\cite{Maaten08dimensionalityreduction,dimension-reduction-a-guided-tour-2}. SNE~\cite{SNE2002}, its variant t-SNE~\cite{van2008visualizing}, and Elastic Embedding~\cite{carreira2010elastic} are among the most successful approaches. To efficiently generate the embedding of out-of-sample data, SNE and t-SNE were, respectively, extended to take a parametric embedding form of a shallow neural network~\cite{MinThesis} and a deep neural network~\cite{Maaten09}. As is discussed in the introduction, the objective functions of neighbor embedding methods have $O(n^2)$ computational complexity for $n$ data points, which limits their applicability only to small datasets. Recently, with the growing importance of big data analytics, several research efforts have been devoted to enhancing the scalability of nonparametric neighbor embedding methods~\cite{van2013barnes,van2014accelerating,yang2013scalable,vladymyrov2014linear}. These methods mainly borrowed ideas from efficient approximations developed for N-body force calculations based on Barnes-Hut trees~\cite{van2013barnes} or fast multipole methods~\cite{greengard1987fast}. Iterative methods with auxiliary variables and second-order methods have been developed to optimize the objective functions of neighbor embedding approaches~\cite{vladymyrov2012partial,vladymyrov2014linear,carreira2015fast}. Particularly, the alternating optimization method with auxiliary variables was shown to achieve faster convergence than mini-batch based conjugate gradient method for optimizing the objective function of pt-SNE. All these scalability handling and optimization research efforts are orthogonal to our development in this paper, because all these methods are designed for the embedding approaches modeling the neighboring relationship between pairwise data points. Therefore, they still have the sensitivity and instability issues, and we can readily borrow these speedup methods to further accelerate our approaches modeling the relationship between data points and exemplars. 

Our proposed method hot-SEE learns a shallow parametric embedding function by considering high-order feature interactions. High-order feature interactions have been studied for learning Boltzmann Machines, autoencoders, structured outputs, feature selection, and biological sequence classification~\cite{DBLP:conf/iccv/Memisevic11,DBLP:conf/aistats/MinNCG14,MinPSB14,DBLP:conf/cvpr/RanzatoH10,DBLP:journals/jmlr/RanzatoKH10,DBLP:journals/corr/GuoZM15,Min2014kdd,MinBio2015,MinGS17}. To the best of our knowledge, our work here is the first successful one to model input high-order feature interactions for unsupervised data embedding and visualization.

Our work in this paper is also related to a recent supervised data embedding method called en-HOPE~\cite{MinGS17}. Unlike en-HOPE, our proposed methods here are unsupervised and have a completely different objective function with different motivations.

\section{Methods}\label{sec:method}
In this section, we introduce the objective of pt-SNE at first. Then we describe the parametric embedding functions of our methods based on a deep neural network as in pt-SNE and a shallow neural network with high-order feature interactions. Finally, we present our proposed parametric stochastic exemplar-centered embedding methods dt-SEE and hot-SEE  with low computational cost.

\subsection{Parametric t-SNE using a Deep Neural Network and a Shallow High-order Neural Network}\label{sec:sup}
Given a set of data points $\mathcal{D} = \{{\mathbf x}^{(i)}: i = 1,\ldots, n\}$, where ${\mathbf x}^{(i)} \in {\mathbb R}^H$ is the input feature vector.
pt-SNE  learns a deep neural network as a nonlinear feature transformation from the high-dimensional input feature space to a low-dimensional latent embedding space 
$\{f({\mathbf x}^{(i)}): i = 1,\ldots, n\}$, where $f({\mathbf x}^{(i)}) \in {\mathbb R}^h$,  and $h < H$.  For data visualization, we set $h = 2$.

pt-SNE assumes, $p_{j|i}$, the probability of each data point $i$ chooses every other data point $j$ as its nearest neighbor in the high-dimensional space follows a Gaussian distribution. The joint probabilities measuring the pairwise similarities between data points ${\mathbf x}^{(i)}$ and ${\mathbf x}^{(j)}$ are defined by symmetrizing two conditional probabilities, $p_{j|i}$ and $p_{i|j}$, as follows,
\begin{eqnarray}
p_{j|i} & = & \frac{\exp(-||{\mathbf x}^{(i)} - {\mathbf x}^{(j)} ||^2/2\sigma_i^2)} {\sum_{k \neq i}\exp(-||{\mathbf x}^{(i)} - {\mathbf x}^{(k)} ||^2/2\sigma_i^2)}, \label{eqn:symmp} \\
p_{i|i} & = & 0, \\
p_{ij} & = & \frac{p_{j|i} + p_{i|j}}{2n}, 
\end{eqnarray}

where the variance of the Gaussian distribution, $\sigma_i$, is set such that the perplexity of the conditional distribution $P_i$ equals a user-specified perplexity $u$ that can be interpreted as the expected number of nearest neighbors of data point $i$. With the same $u$ set for all data points, $\sigma_i$'s tend to be smaller in regions of higher data densities than the ones in regions of lower data densities. The optimal value of $\sigma_i$ for each data point $i$ can be easily found by a simple binary search~\cite{SNE2002}. Although the user-specified perplexity $u$ makes the variance $\sigma_i$ for each data point $i$ adaptive, the embedding performance is still very sensitive to this hyperparameter, which will be discussed later. In the low-dimensional space, pt-SNE assumes, the neighboring probability between pairwise data points $i$ and $j$,  $q_{ij}$, follows a heavy-tailed student t-distribution. The student t-distribution is able to, on one hand, measure the similarities between pairwise low-dimensional points, on the other hand, allow dissimilar objects to be modeled far apart in the embedding space, avoiding crowding problems.  
 \begin{eqnarray}
q_{ij} & = & \frac{(1 + ||f({\mathbf x}^{(i)}) -  f({\mathbf x}^{(j)})||^2)^{-1}} {\sum_{kl:k \neq l}  (1 + ||f({\mathbf x}^{(k)}) -  f({\mathbf x}^{(l)})||^2)^{-1}},  \label{eqn:symmq} \\
q_{ii} & = & 0. 
\end{eqnarray}

To learn the parameters of the deep embedding function ${\mathbf f}(.)$, pt-SNE minimizes the following Kullback-Leibler divergence between the joint distributions $P$ and $Q$ using Conjugate Gradient descent,
\begin{equation}\label{obj}
\small
\ell = KL (P || Q) = \sum_{ij: i \neq j} p_{ij}\log \frac{p_{ij}}{q_{ij}}.
\end{equation}
The above objective function has $O(n^2)$ terms defined over pairwise data points, which is computationally prohibitive and prevents pt-SNE from scaling to a fairly big dataset. To overcome such scalability issue, heuristic mini-batch approximation is often used in practice. However, as will be shown in our experiments, pt-SNE is unstable and highly sensitive to the chosen batch size to achieve reasonable performance. This is due to the dilemma of the quadratic cost function approximation and DNN optimization through mini-batches: approaching the true objective requires large batch sizes but finding a good local minimum benefits from small batch sizes.

Although pt-SNE based on a deep neural network has a powerful nonlinear feature transformation, parameter learning is hard and requires complicated procedures such as tuning network architectures and tuning many hyperparameters. Most users who are only interested in data embedding and visualization are reluctant to go through these complicated procedures. Here we propose to use high-order feature interactions, which often capture structural knowledge of input data, to learn a shallow parametric embedding model instead of a deep model. The shallow model is much easier to train and does not have many hyperparameters. In the following, the shallow high-order parametric embedding function will be presented. We expand each input feature vector ${\mathbf x}$ to have an additional component of $1$ for absorbing bias terms, that is, ${\mathbf x}^{\prime} = [{\mathbf x};   1]$, where ${\mathbf x}^{\prime} \in {\mathbb R}^{H+1}$. The $O$-order feature interaction is the product of all possible $O$ features $\{x_{i_1}\times \ldots \times  x_{i_t} \times \ldots \times x_{i_O}\}$ where,  $t \in \{1, \ldots, O\}$, and $\{i_1, \ldots, i_t, \ldots, i_O\} \in \{1, \ldots, H\}$. Ideally, we want to use each $O$-order feature interaction as a coordinate and then learn a linear transformation to map all these high-order feature interactions to a low-dimensional embedding space. However, it's very expensive to enumerate all possible $O$-order feature interactions. For example, if $H = 1000, O = 3$, we must deal with a $10^9$-dimensional vector of high-order features. We approximate a Sigmoid-transformed high-order feature mapping ${\mathbf y} = f({\mathbf x})$ by constrained tensor factorization as follows, 
\begin{equation}
\label{shopemap}
y_s = \sum_{k=1}^m V_{sk} \sigma(\sum_{f=1}^F W_{fk}({\mathbf C_f}^T {\mathbf x}^{\prime})^O + b_k),
\end{equation}
where $b_k$ is a bias term, ${\mathbf C}  \in {\mathbb R}^{(H+1) \times F}$ is a factorization matrix, ${\mathbf C}_f$ is the $f$-th column of ${\mathbf C}$, ${\mathbf W}\in {\mathbb R}^{F\times m}$ and ${\mathbf V}\in {\mathbb R}^{h\times m}$ are projection matrices, $y_s$ is the $s$-th component of ${\mathbf y}$, $F$ is the number of factors, $m$ is the number of high-order hidden units, and $\sigma (x) = \frac{1}{1 + e^{-x}}$.  Because the last component of $\mathbf{x}^\prime$ is 1 for absorbing bias terms, the full polynomial expansion of $({\mathbf C_f}^T {\mathbf x}^\prime)^O$ essentially captures all orders of input feature interactions up to order $O$. Empirically, we find that $O=2$ works best for all datasets we have and set $O=2$ for all our experiments. The hyperparameters $F$ and $m$ are set by users. Combining Equation~\ref{obj}, Equation~\ref{eqn:symmp}, Equation~\ref{eqn:symmq} and the feature transformation function in Equation~\ref{shopemap} leads to a method called high-order t-SNE (hot-SNE). As pt-SNE, the objective function of hot-SNE involves comparing pairwise data points and thus has quadratic computational complexity with respect to the sample size. The parameters of hot-SNE are learned by Conjugate Gradient descent as in pt-SNE. 

\subsection{Parametric t-Distributed Stochastic Exemplar-centered Embedding}
To address the instability, sensitivity, and unscalability issues of pt-SNE, we present deep t-distributed stochastic exemplar-centered embedding (dt-SEE) and high-order t-distributed stochastic exemplar-centered embedding (hot-SEE) building upon pt-SNE and hot-SNE for parametric data embedding described earlier. The resulting objective function has significantly reduced computational complexity with respect to the size of training set compared to pt-SNE. The underlying intuition is that, instead of comparing pairwise training data points, we compare training data only with a small number of representative exemplars in the training set for neighborhood probability computations. To this end, we simply precompute the exemplars by running a fixed number of iterations of k-means with scalable k-means++ seeding on the training set, which has at most linear computational complexity with respect to the size of training set~\cite{bahmani2012scalable}.

Formally, given the same dataset $\mathcal{D}$ with formal descriptions as introduced in Section~\ref{sec:sup}, 
we perform a fixed number of iterations of k-means updates on the training data to identify $z$ exemplars from the whole dataset, where $z$ is a user-specified free parameter and $z << n$ (please note that k-means often converges within a dozen iterations and shows linear computational cost in practice).  Before performing k-means updates, the exemplars are carefully seeded by scalable k-means++, which will make our methods robust under abnormal conditions, although our experiments show that random seeding works equally well. We denote these exemplars by $\{{\mathbf e}^{(j)}:  j = 1, \ldots, z\}$.  The high-dimensional neighboring probabilities is calculated through a Gaussian distribution,
\begin{eqnarray}
p_{j|i} & = & \frac{\exp(-||{\mathbf x}^{(i)} - {\mathbf e}^{(j)} ||^2/2\sigma_i^2)} {\sum_k\exp(-||{\mathbf x}^{(i)} - {\mathbf e}^{(k)} ||^2/2\sigma_i^2)}, \label{eqn:symmp_exem} \\
p_{j|i} & = & \frac{p_{j|i}}{n}, 
\end{eqnarray}
where $i = 1, \ldots, n, j=1, \ldots, z$, and the variance of the Gaussian distribution, $\sigma_i$, is set such that the perplexity of the conditional distribution $P_i$ equals a user-specified perplexity $u$ that can be interpreted as the expected number of nearest exemplars, not neighboring data points, of data instance $i$. Since the high-dimensional exemplars capture both local data density distributions and global clustering patterns, different choices of perplexities over exemplars will not change the embedding too much, resulting in much more robust visualization performance than that of other embedding methods insisting on modeling local pairwise neighboring probabilities.

Similarly, the low-dimensional neighboring probabilities is calculated through a t-distribution,
\begin{eqnarray}
q_{j|i} & = & \frac{(1 + d_{ij})^{-1}} {\sum_{i=1}^n\sum_{k=1}^z (1 + d_{ik})^{-1}}, \label{eqn:q_exem}\\
d_{ij} & = & ||f({\mathbf x}^{(i)}) - f({\mathbf e}^{(j)}) ||^2,
\end{eqnarray}
where $f(\cdot)$ denotes a deep neural network for dt-SEE or the high-order embedding function as described in Equation~\ref{shopemap} for hot-SEE. 

Then we minimize the following objective function to learn the embedding parameters ${\mathbf \Theta}$ of dt-SEE and hot-SEE while keeping the exemplars $\{{\mathbf e}^{(j)}\}$ fixed,
\begin{eqnarray}\label{exobj}
&\min_{}^{} \ell({\mathbf \Theta}, \{{\mathbf e}^{(j)}\})  =  \sum_{i=1}^{n}\sum_{j=1}^{z} p_{j|i}\log \frac{p_{j|i}}{q_{j|i}}
\end{eqnarray}
where $i$ indexes training data points, $j$ indexes exemplars, ${\mathbf \Theta}$ denotes the high-order embedding parameters $\{\{b_k\}_{k=1}^m, \mathbf{C }, \mathbf{W}, \mathbf{V}\}$ in Equation~\ref{shopemap}.

Note that unlike the probability distribution in Equation~\ref{eqn:symmq},  $q_{j|i}$ here is computed only using the pairwise distances between training data points and exemplars. This small modification has significant benefits. Because $z << n$, compared to the quadratic computational complexity with respect to $n$ of Equation~\ref{obj}, the objective function in Equation~\ref{exobj} has a significantly reduced computational cost, considering that the number of representative exemplars is often much much smaller than $n$ for real-world large datasets in practice. 

\subsection{Further Reduction on Computational Complexity and Memory Complexity by Noise Contrastive Estimation}
We can even further reduce the computational complexity and memory complexity of dt-SEE and hot-SEE using noise contrastive estimation (NCE). Instead of computing neighboring probabilities between each data point $i$ and all $z$ exemplars, we can simply only compute the probabilities between data point $i$ and its $z_e$ nearest exemplars for both $P$ and $Q$. For high-dimensional probability distribution $P_i$, we simply set the probabilities between $i$ and other exemplars 0; for low-dimensional probability distribution $Q_i$, we randomly sample $z_n$ non-neighboring exemplars outside of these $z_e$ neighboring exemplars, and use the sum of these $z_n$ non-neighboring probabilities multiplied by a constant $K_e$ and the $z_e$ neighboring probabilities to approximate the normalization terms involving data point $i$ in Equation~\ref{eqn:q_exem}. Since this strategy based on noise contrastive estimation eliminates the need of computing neighboring probabilities between data points and all exemplars, it further reduces computational and memory complexity.

\section{Experiments}\label{sec:experiment}
In this section, we evaluate the effectiveness of dt-SEE and hot-SEE by comparing them against state-of-the-art unsupervised parametric embedding method pt-SNE based upon three datasets, \textit{i.e.}, COIL100, MNIST, and Fashion.  The COIL100 data~\footnote{http://www1.cs.columbia.edu/CAVE/software/softlib/coil-100.php} contains 7200 images with 100 classes, where 3600 samples for training and 3600 for test. The MNIST dataset~\footnote{http://yann.lecun.com/exdb/mnist/} consists of 60,000 training and 10,000 test gray-level 784-dimensional images. The Fashion dataset~\footnote{https://github.com/zalandoresearch/fashion-mnist} has the same number of classes, training and test data points as that of MNIST, but is designed to classify 10 fashion  products, such as boot, coat, and bag, where each contains a set of pictures taken by professional photographers from different aspects of the product, such as looks from front, back, with model, and in an outfit.

To make computational procedures and tuning procedures for data visualization simpler, none of these models was pre-trained using any unsupervised learning strategy, although hot-SNE, hot-SEE, dt-SEE, and pt-SNE could all be pre-trained by autoencoders or variants of Restricted Boltzmann Machines~\cite{MinMYBZ10,MinBio2015}. 

For hot-SNE and hot-SEE, we set $F=800$ and $m=400$ for all the datasets used. For pt-SNE and dt-SEE, we set the deep neural network architecture to input dimensionality $H$-500-500-2000-2 for all datasets, following the architecture design in van der Maaten (2009)~\cite{Maaten09}. For hot-SEE and dt-SEE, when the exemplar size is smaller than 1000, we set batch size to 100; otherwise, we set it 1000. With the above architecture design, the shallow high-order neural network used in hot-SNE and hot-SEE is as fast as $2.5$ times of the deep neural network used in pt-SNE and dt-SEE for embedding $10,000$ MNIST test data.

For all the experiments, the predictive accuracies were obtained by the 1NN approach on top of the 2-dimensional representations generated by different methods. The error rate was calculated by the number of misclassified test data points divided by the total number of test data points. 

\subsection{Performance Comparisons with Different Batch Sizes and Perplexities on COIL100 and MNIST}
\begin{figure}[H]
      \subfloat[COIL100\label{COIL100}]{%
      \includegraphics[width=0.4738\textwidth]{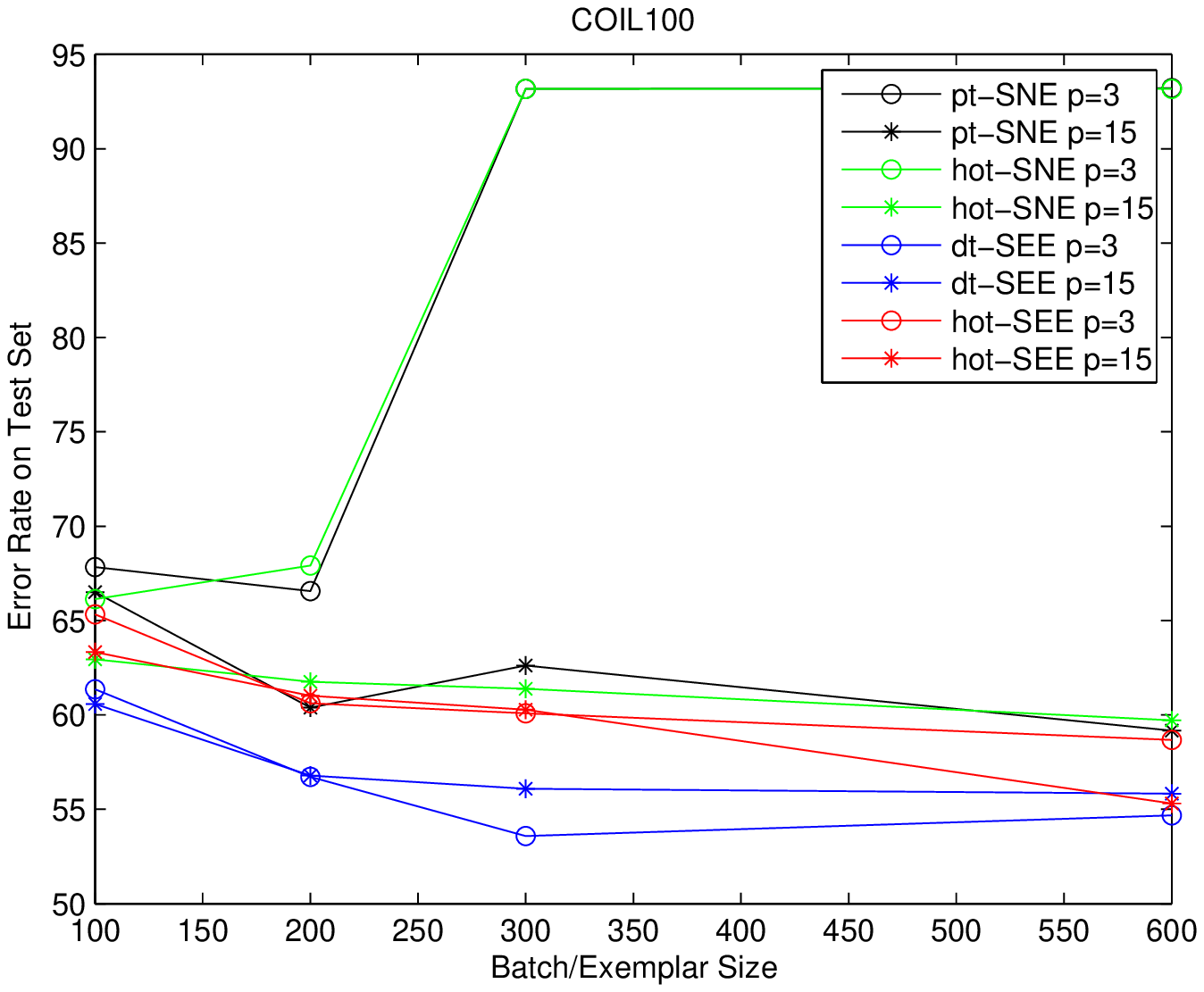}
      }
      \hfill
        \subfloat[MNIST\label{MNISTd}]{%
      \includegraphics[width=0.4738\textwidth]{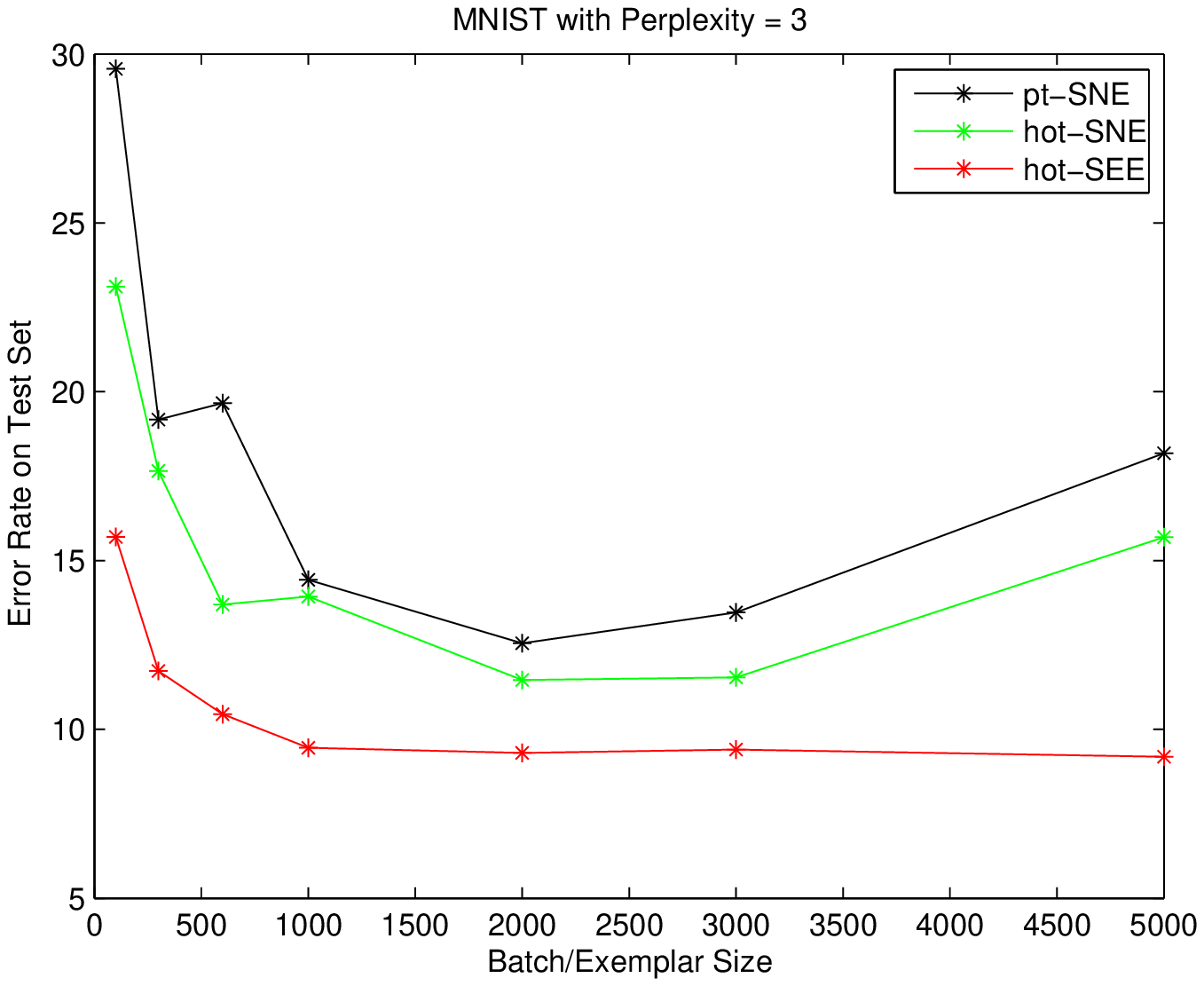}
    }
  \centering
  \caption{batch size sensitivity test on COIL100 and MNIST }
  \label{fig:batchsizeSensitivity}
  \end{figure}

\begin{figure}[H]
      \subfloat[COIL100\label{COIL100}]{%
      \includegraphics[width=0.4738\textwidth]{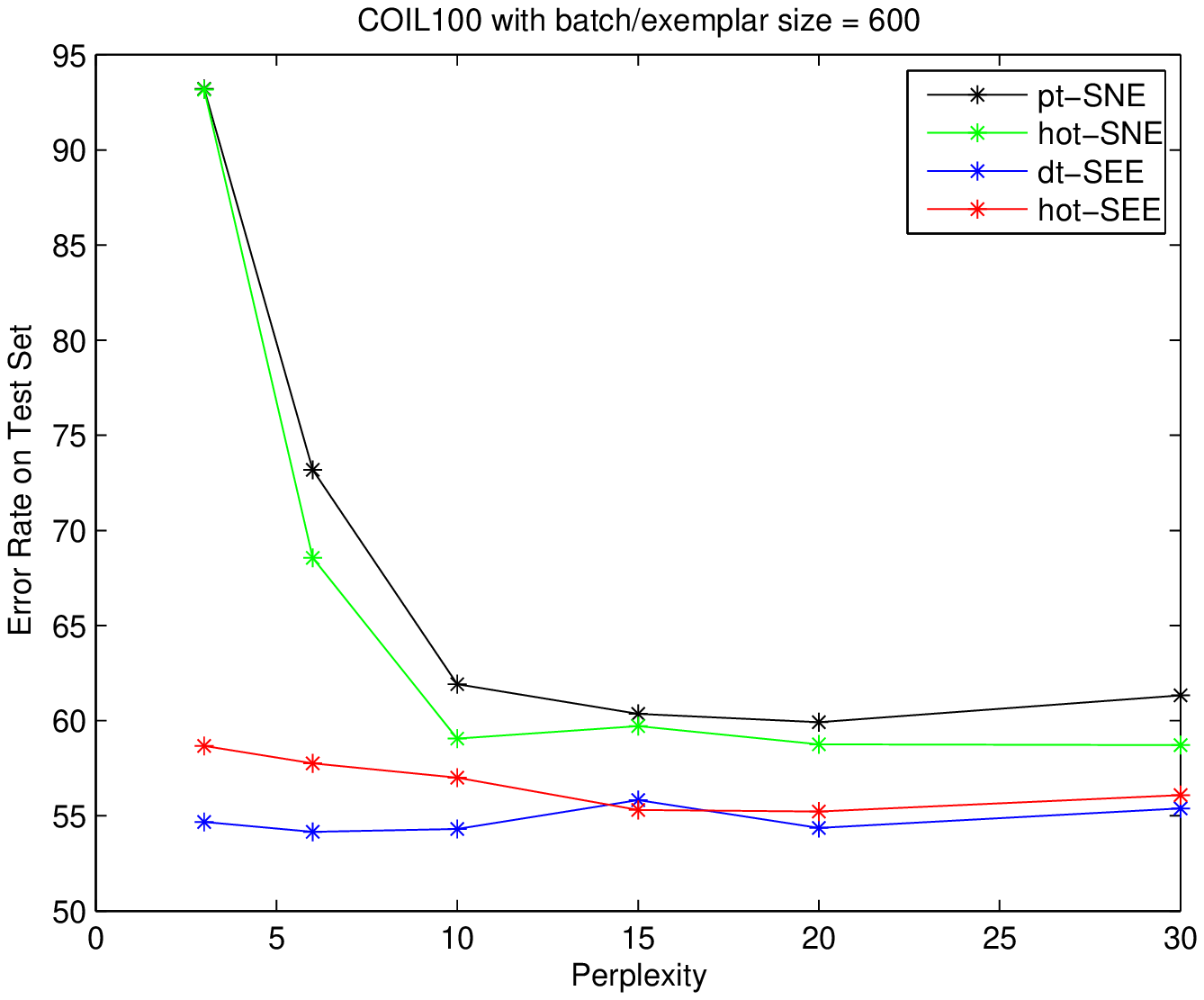}
      }
      \hfill
        \subfloat[MNIST\label{MNISTd}]{%
      \includegraphics[width=0.4738\textwidth]{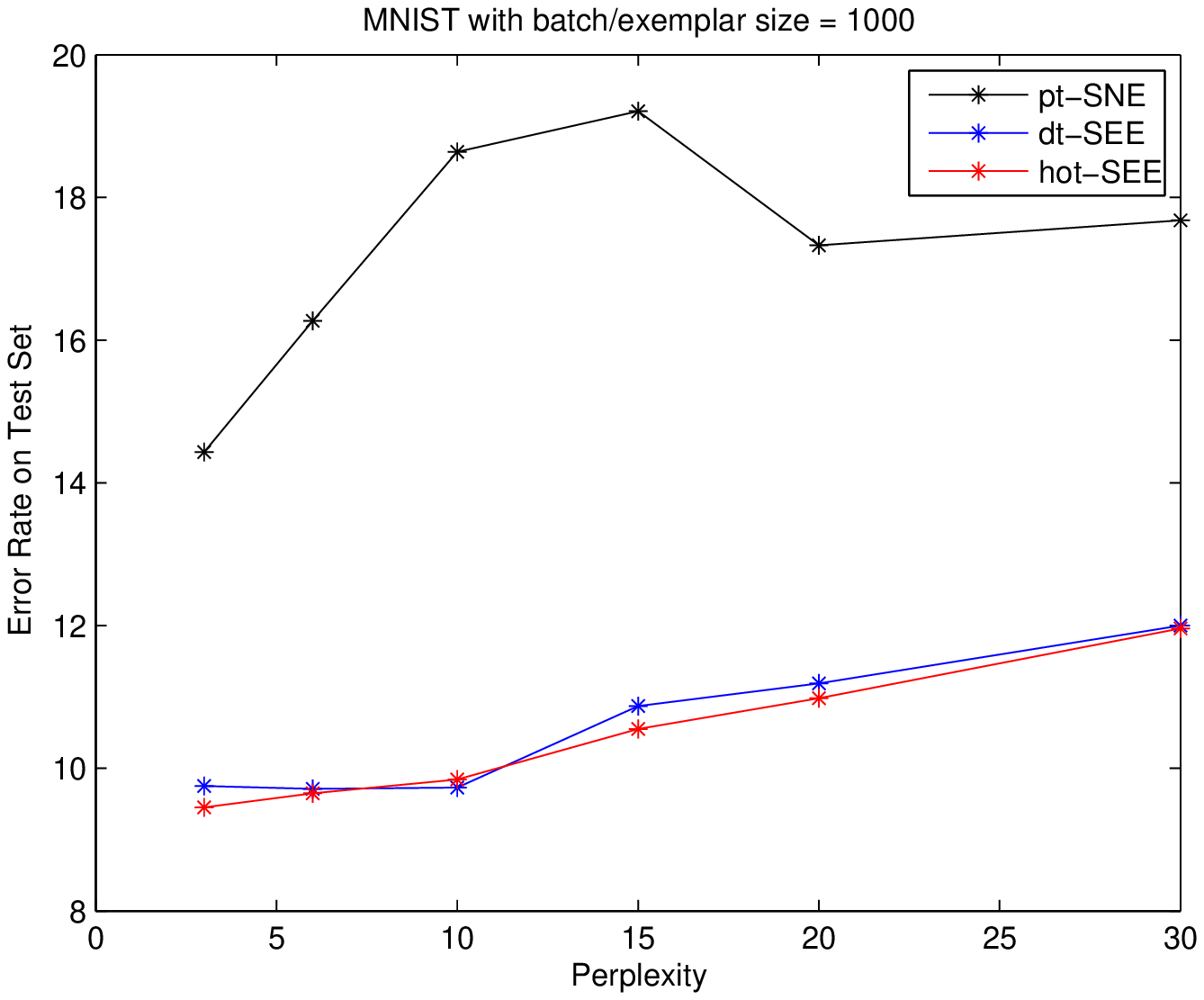}
    }
  \centering
  \caption{perplexity sensitivity test on COIL100 and MNIST }
  \label{fig:perplexitySensitivity}
  \end{figure}

Our first experiment aims at examining the robustness of different testing methods with respect to  the batch size and the perplexity used. Figures~\ref{fig:batchsizeSensitivity} and~\ref{fig:perplexitySensitivity}  depict our results on the COIL100 and MNIST datasets when varying the batch size and perplexity, respectively, used by the testing methods.

Figure~\ref{fig:batchsizeSensitivity} suggests that, for the COIL100 data, the pt-SNE was very sensitive to the selection of  the batch size; efforts were needed to find a right batch size in order to obtain good performance. On the other hand, the use of different batch sizes had very minor impact on the predictive performance of both the dt-SEE and hot-SEE strategies. Similarly, for the MNIST data, as shown in Figure~\ref{fig:perplexitySensitivity}, in order to obtain good predictive performance, the pt-SNE needed to have a batch size not too big and not too small. On the contrary, the hot-SEE methods was insensitive to the size of batch larger than 300. 

Based on the results in Figure~\ref{fig:batchsizeSensitivity}, we selected the best batch sizes for both the COIL100 and MNIST data sets, with 600 and 1000, respectively, but we varied the values of the perplexities used. In Figure~\ref{fig:perplexitySensitivity}, one can observe that, the performance of the pt-SNE and hot-SNE could dramatically change due to the use of different perplexities, but that was not the case for both the dt-SEE and hot-SEE. Similarly, for the MNIST data, as depicted in Figure~\ref{fig:perplexitySensitivity}, in order to obtain good predictive performance, one would need to carefully tune for the right perplexity.  On the contrary, both the dt-SEE and hot-SEE methods performed quite robust with respect to different selected perplexities.

Because the choices of batch size and perplexity are coupled in a complicated way in pt-SNE as explained in the introduction, we run additonal experiments to show the advantages of dt-see and hot-see. When we set perplexity to 10 and batch size to 100, 300, 600, 1000, 2000, 3000, 5000, 10000, the test error rate of pt-SNE on MNIST is, respectively, $32.97\%$, $22.1\%$, $24.00\%$, $16.30\%$, $12.41\%$, $12.28\%$, $13.09\%$, $16.43\%$, which still varies a lot. In contrast, the error rates of dt-SEE or hot-SEE using 1000 exemplars are consistently below $10\%$ with the same batch size ranging from 100 to 10000 and perplexity 3 and 10, which again shows shat exemplar-centered embedding dt-see and hot-see are much more robust than pt-SNE.

\subsection{Experimental Results on the Fashion dataset}
We also further evaluated the predictive performance of the testing methods using the Fashion data set. We used batch sizes of 1000 and 2000, along with perplexity of 3 in all the experiments since both pt-SNE and hot-SNE favored these settings as suggested in Figures~\ref{fig:batchsizeSensitivity} and~\ref{fig:perplexitySensitivity}. The achieved accuracies are  shown in Table~\ref{tab:vggknn}.

\begin{table}[H]
\centering
 \begin{tabular}{|c|c|}\hline
 Methods & Error Rates \\
 \hline 
pt-SNE (batchsize = 1000) & 32.48\\
\hline
pt-SNE (batchsize = 2000) & 32.04\\
\hline 
hot-SNE (batchsize = 1000) & 31.29\\
\hline
hot-SNE (batchsize = 2000) & 31.82\\
\hline
dt-SEE (batchsize = 1000) & 29.42\\
\hline
dt-SEE (batchsize = 2000) & 28.30\\
\hline 
hot-SEE (batchsize = 1000) & 29.06\\
\hline
hot-SEE (batchsize = 2000) & \textbf{28.18}\\
\hline
\end{tabular}
\caption{Error rates (\%) by 1NN on the 2-dimensional representations produced by different methods with perplexity = 3 on the Fashion dataset.}\label{tab:vggknn}
\end{table}

Results in Table~\ref{tab:vggknn} further confirmed the superior performance of our methods. Both the dt-SEE and hot-SEE significantly outperformed the pt-SNE and hot-SNE.

\subsection{Two-dimensional Visualization of Embeddings}
This section provides the visual results of the embeddings formed by the pt-SNE and hot-SEE methods. 

\begin{figure*}[h]
      \subfloat[pt-SNE, batch size 100\label{COIL100}]{%
      \includegraphics[width=0.503738\textwidth]{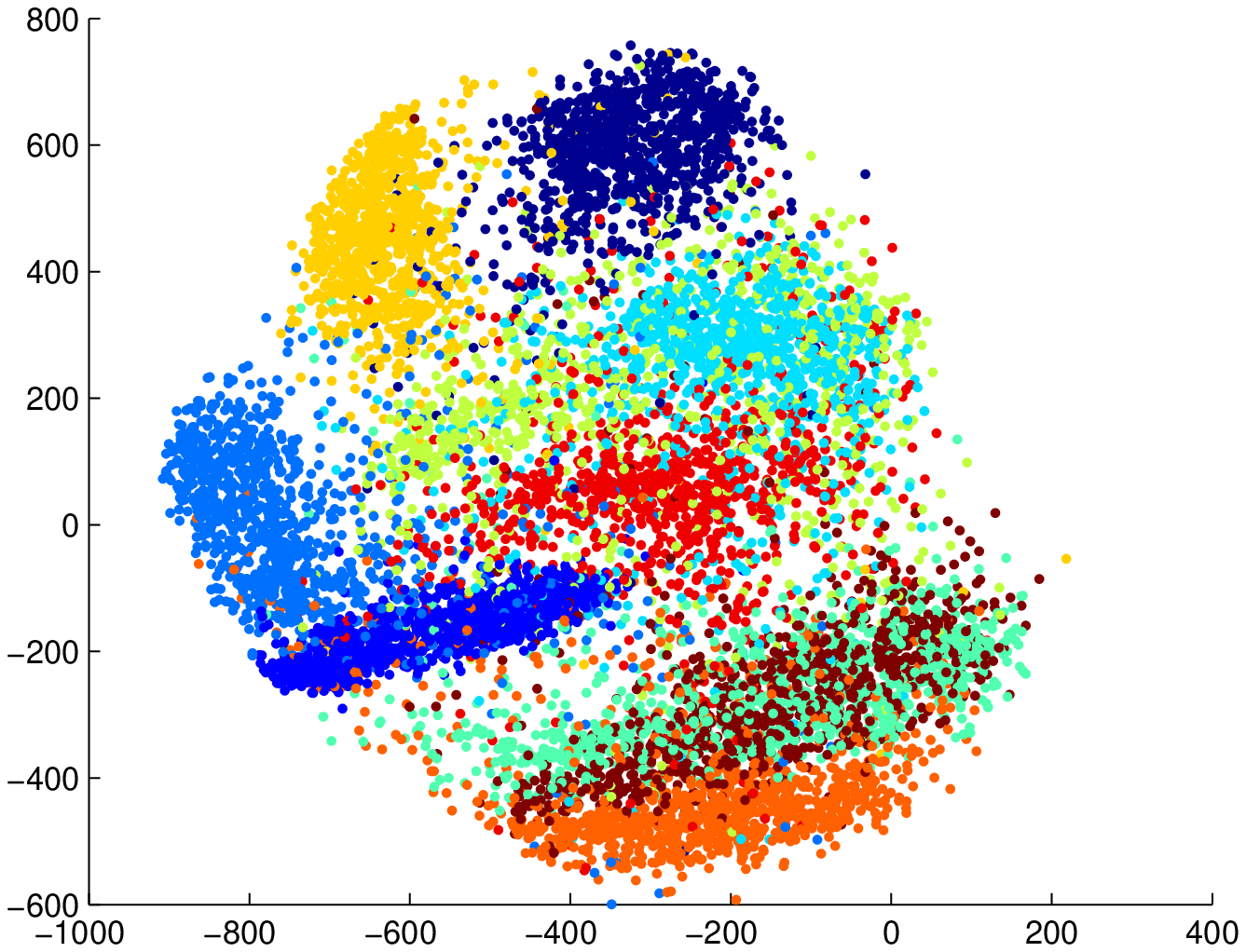}
      }
        \subfloat[hot-SEE, batch size 100\label{MNISTd}]{%
      \includegraphics[width=0.503738\textwidth]{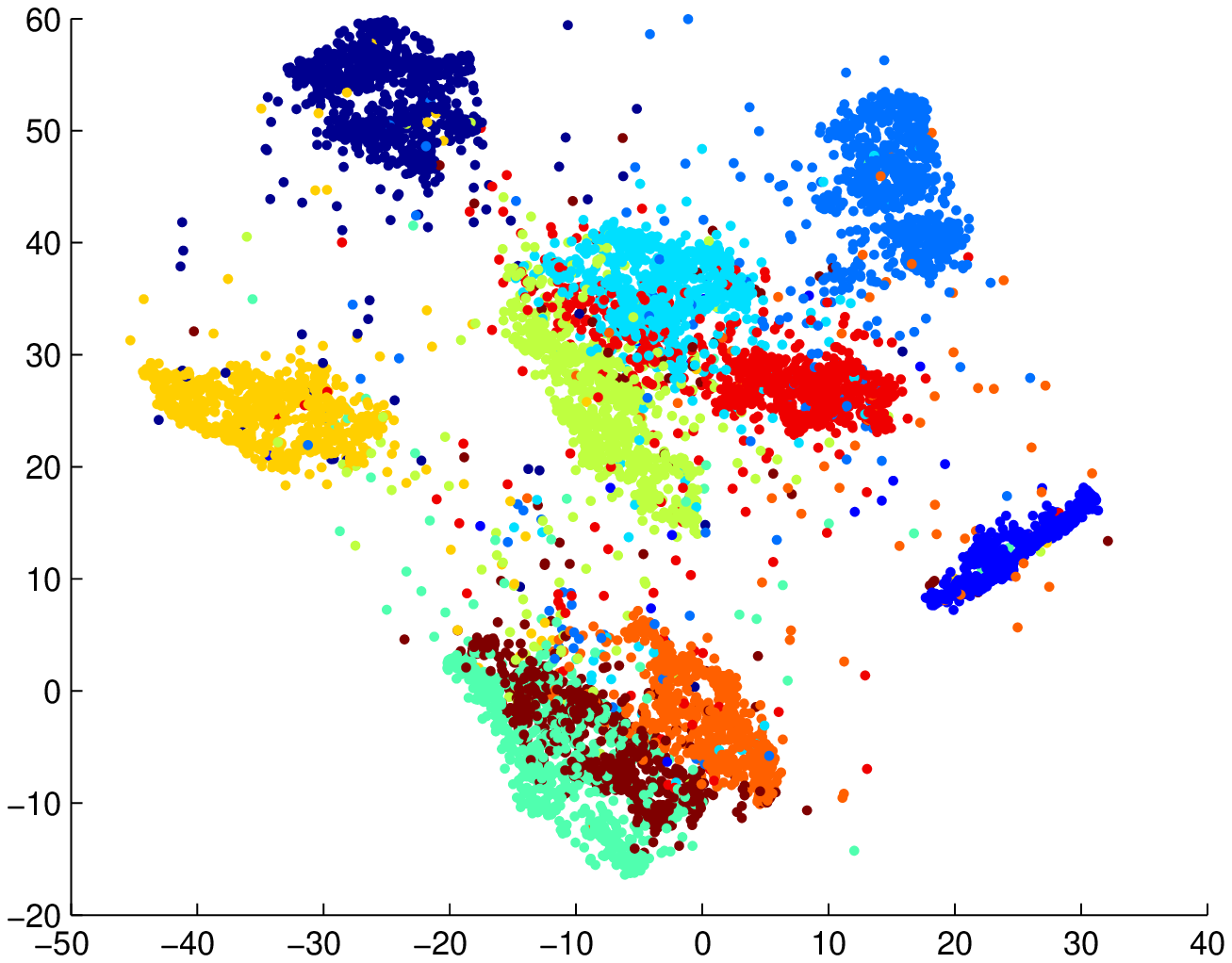}
    }
    		 \hfill
          \subfloat[pt-SNE, perplexity 10\label{COIL100}]{%
      \includegraphics[width=0.503738\textwidth]{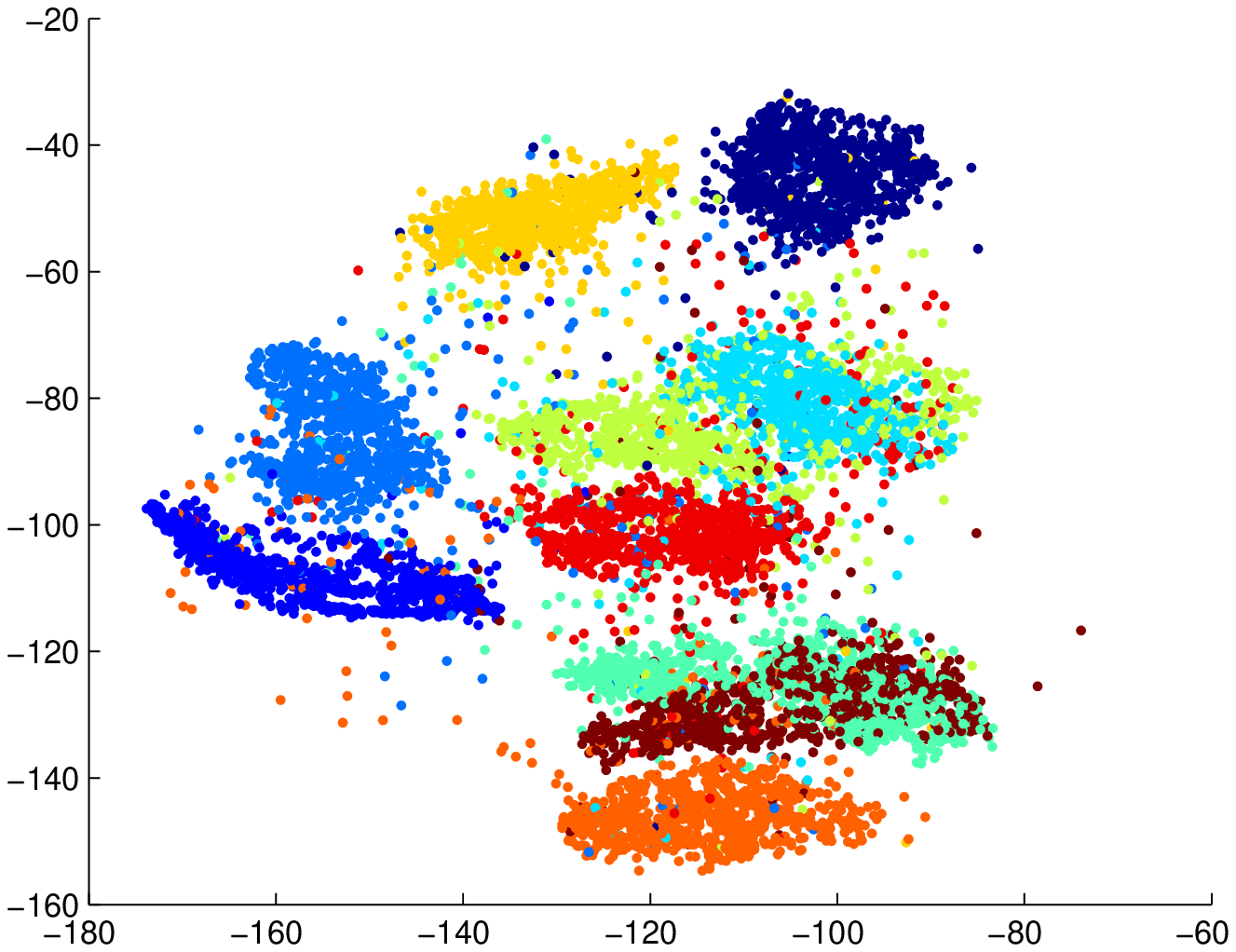}
      }
        \subfloat[hot-SEE, perplexity 10\label{MNISTd}]{%
      \includegraphics[width=0.503738\textwidth]{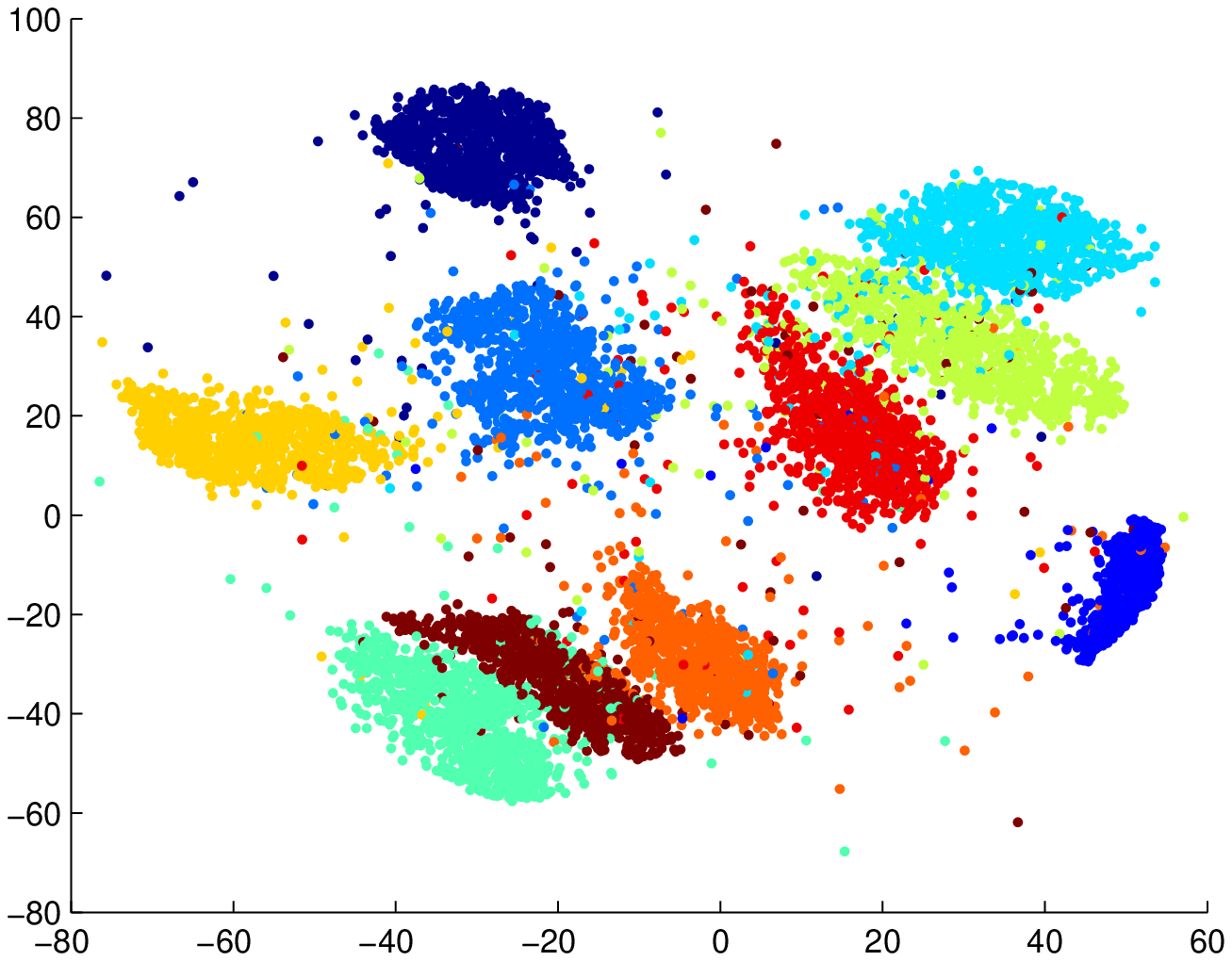}

}
  \centering
  \caption{Comparing pt-SNE to hot-SEE with a small batch size = 100 (perplexity = 3) or a reasonable perplexity = 10 (batch size = 1000) to illustrate pt-SNE's unstable visual performance.}
  \label{fig:ministSmallBatchSize}
  \end{figure*}

  The top and bottom subfigures in Figure~\ref{fig:ministSmallBatchSize} depicts the 2D embeddings on the MNIST data set created by pt-SNE and hot-SEE,  with batch size of 100 (perplexity = 3) and  perplexity of 10 (batch size = 1000), respectively. From these visual figures, one may conclude that the hot-SEE was more stable compared to its competitor pt-SNE. 
	
 \begin{figure*}[h]
      \subfloat[pt-SNE\label{COIL100}]{%
      \includegraphics[width=0.50139738\textwidth]{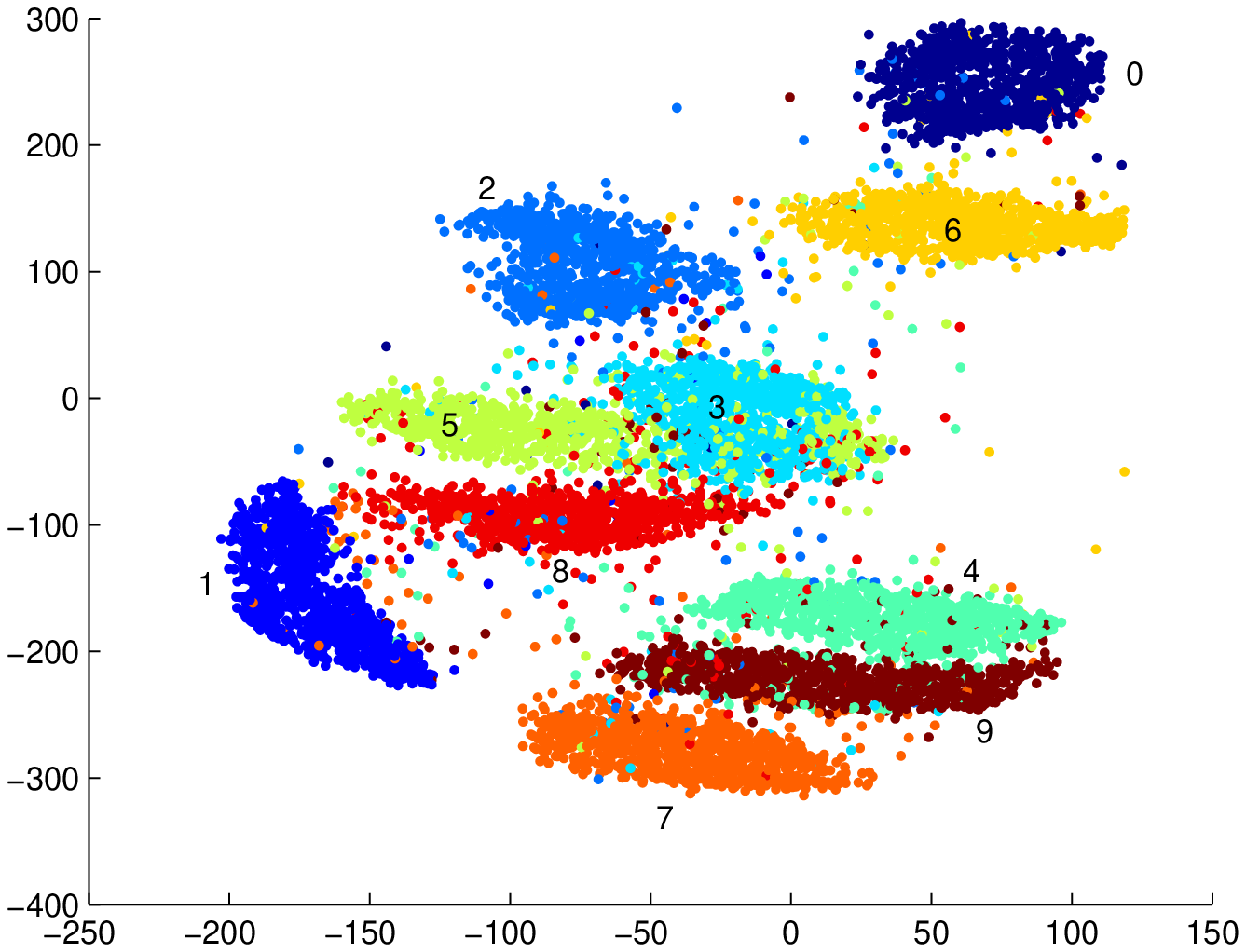}
      }
        \subfloat[hot-SNE\label{MNISTd}]{%
      \includegraphics[width=0.50139738\textwidth]{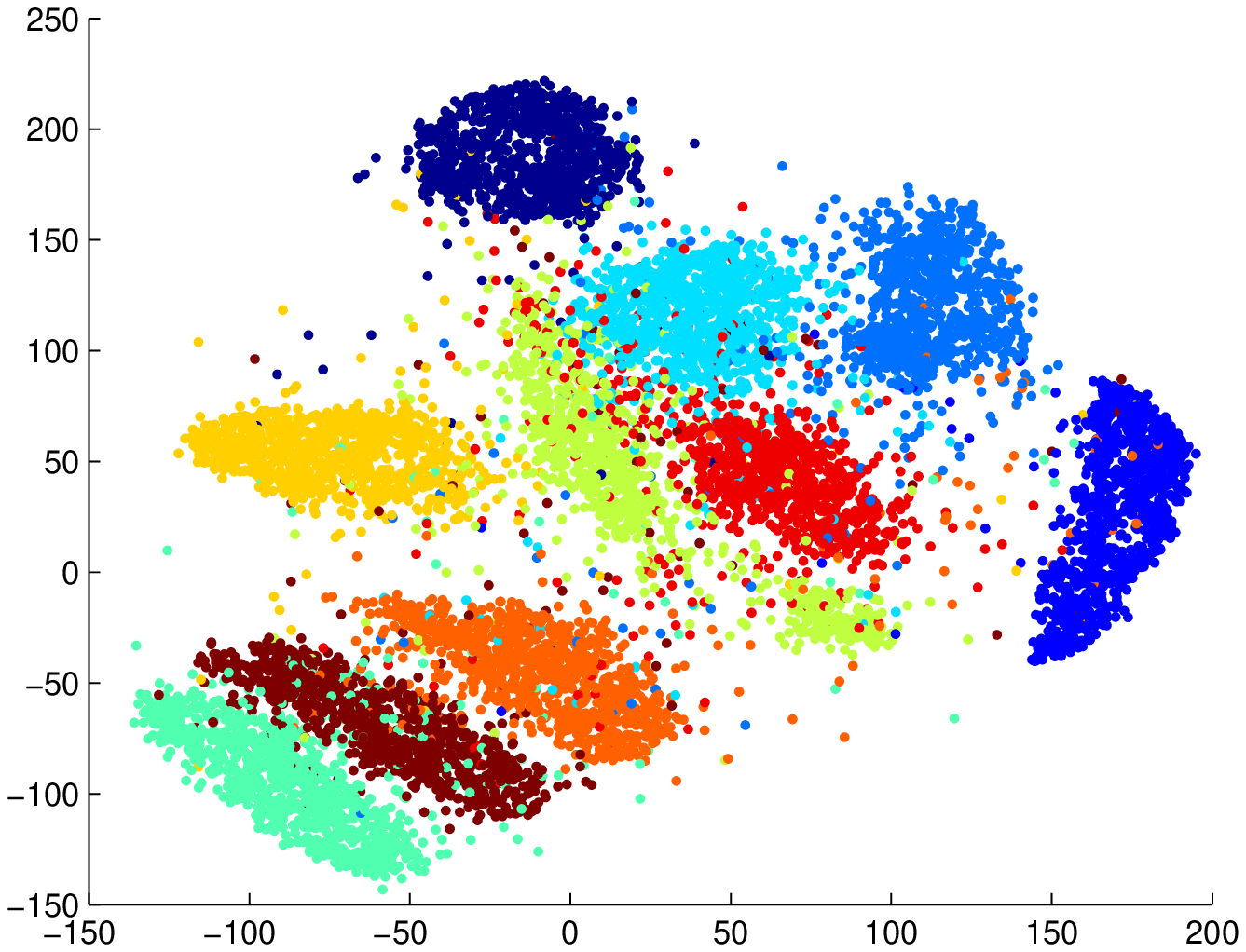}
    }
		 \hfill
      \subfloat[dt-SEE\label{COIL100}]{%
      \includegraphics[width=0.50139738\textwidth]{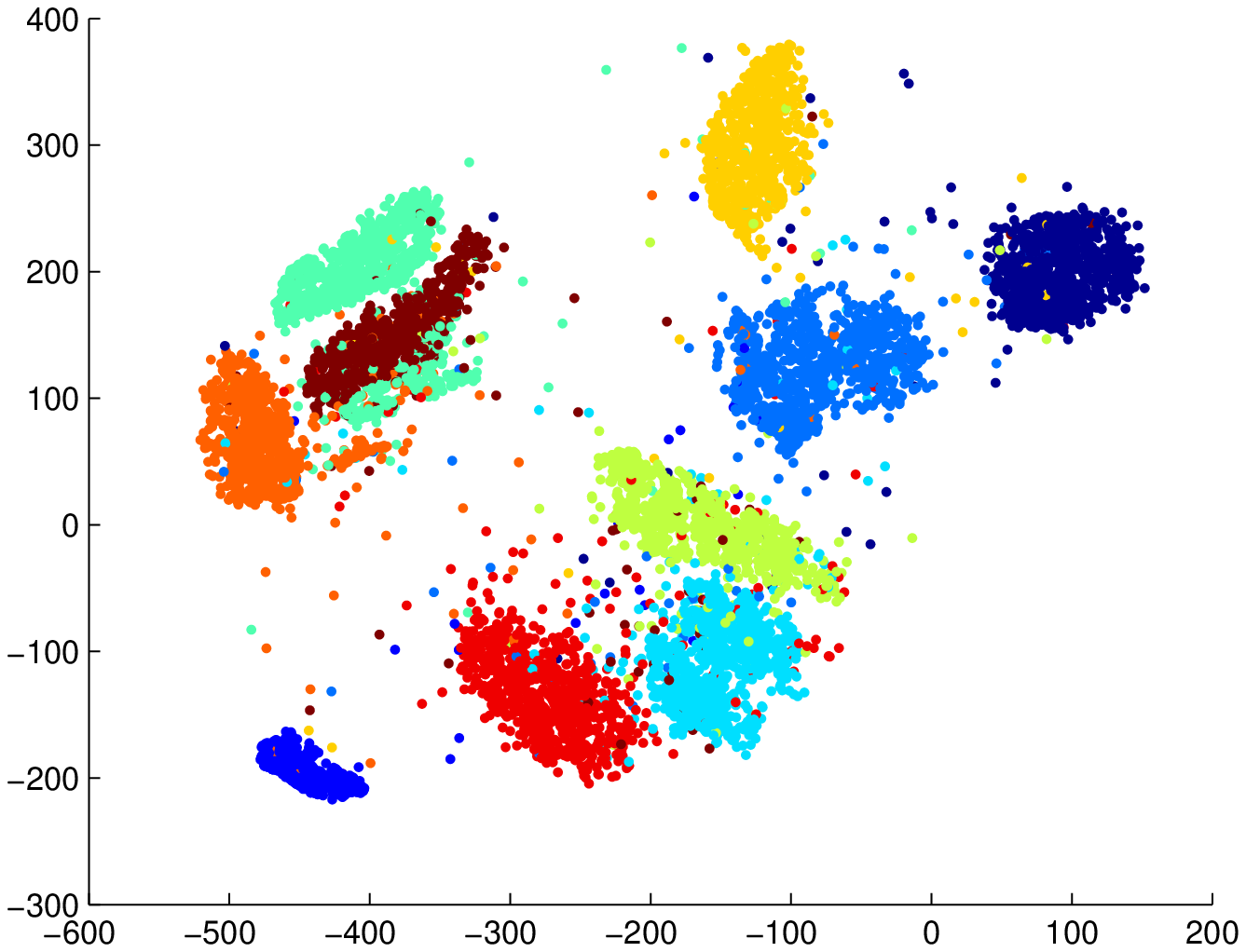}
      }
        \subfloat[hot-SEE\label{MNISTd}]{%
      \includegraphics[width=0.50139738\textwidth]{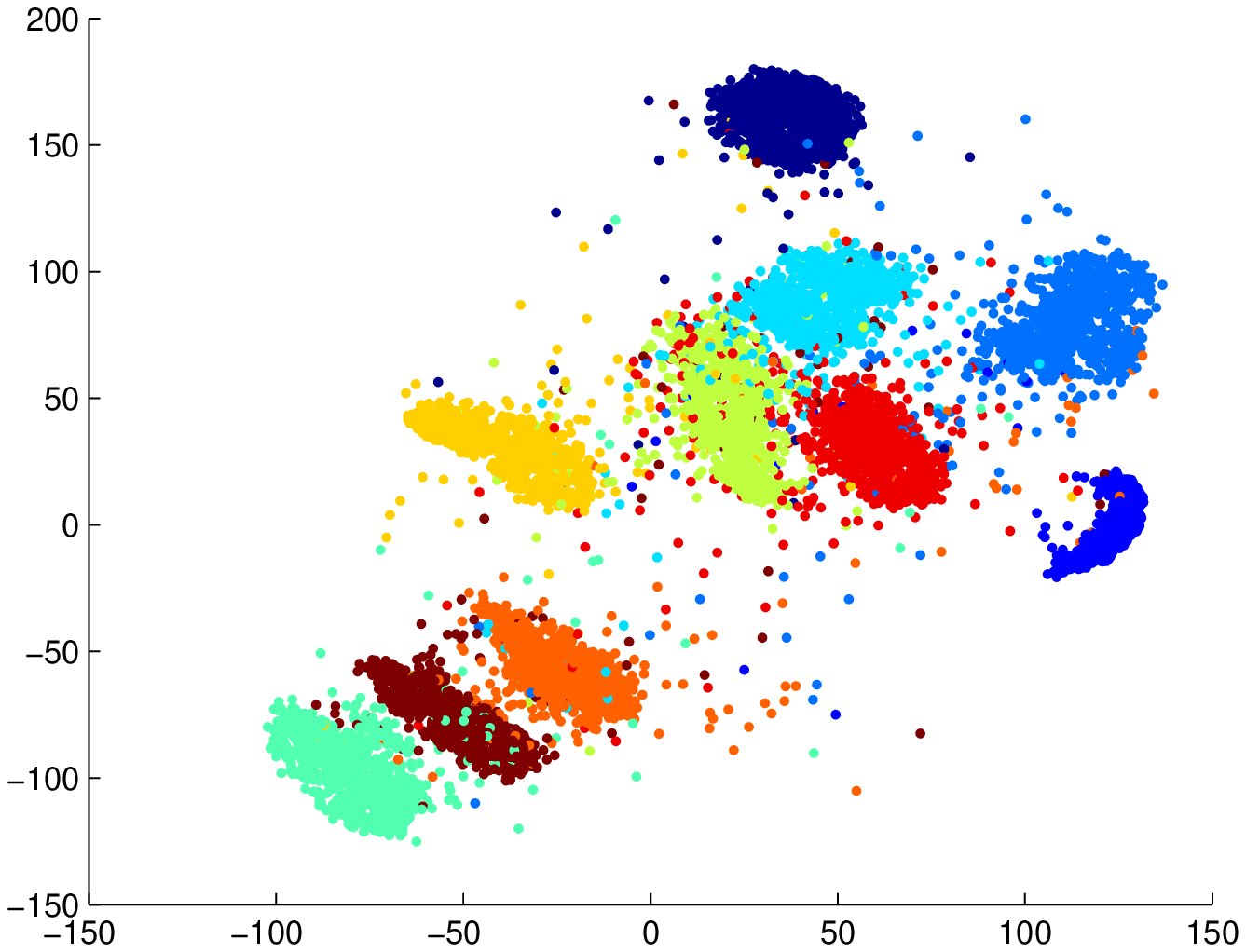}
    }
  \centering
  \caption{MNIST embedding figures for pt-SNE, hot-SNE, dt-SEE, and hot-SEE }
  \label{fig:mnistEmbedding}
  \end{figure*}

  In Figure~\ref{fig:mnistEmbedding}, we also provided the visual results of the MNIST embeddings created by  pt-SNE, hot-SNE, dt-SEE, and hot-SEE, with batch size of 2000.  These results imply that 
 the dt-SEE and hot-SEE produced  the  best visualization:   the data points in each cluster were close to each other but with large separation between different clusters, compared to that of the  pt-SNE and hot-SNE methods.

 \begin{figure}[H]
      \subfloat[pt-SNE\label{COIL100}]{%
      \includegraphics[width=0.483738\textwidth]{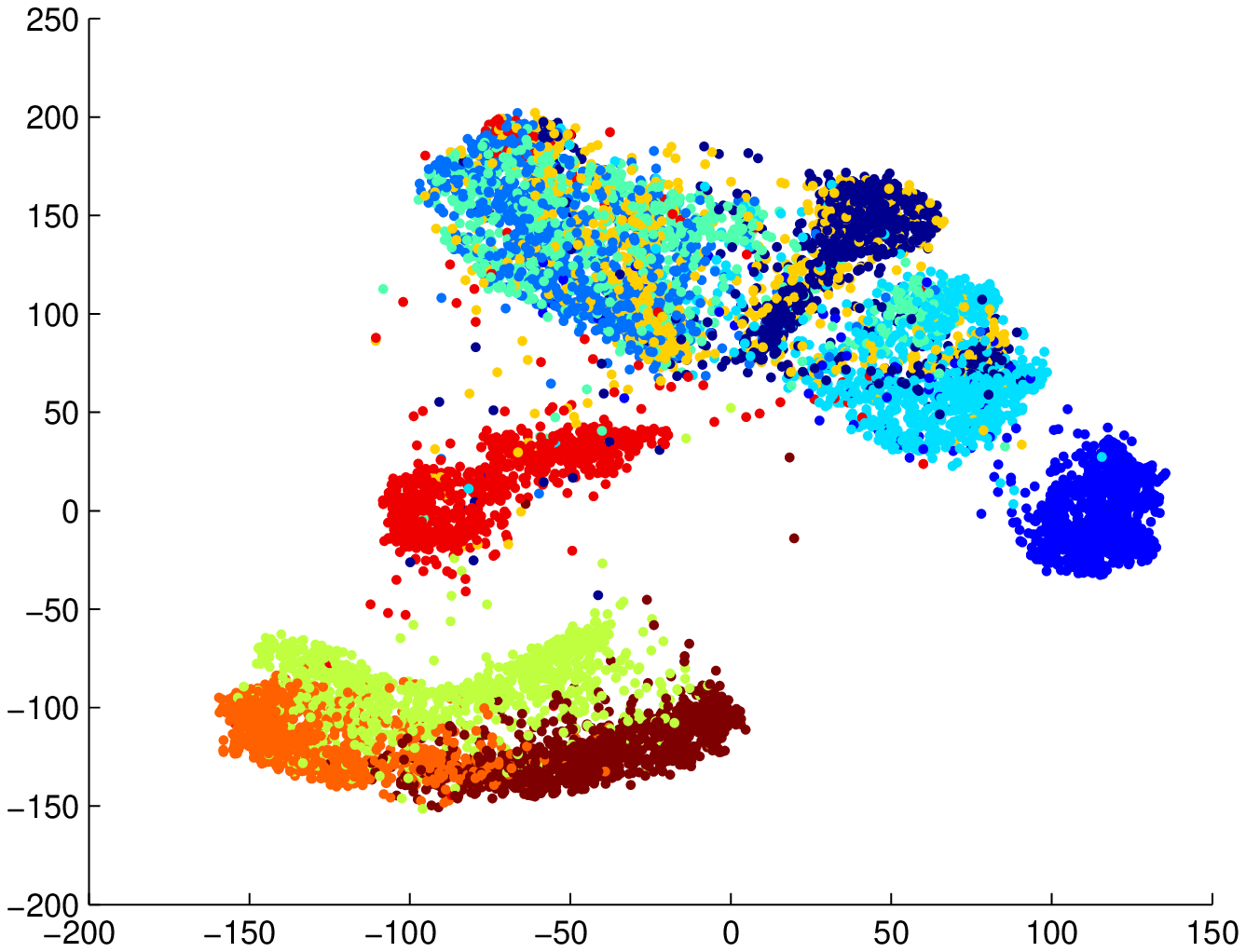}
      }
      \hfill
        \subfloat[hot-SEE\label{MNISTd}]{%
      \includegraphics[width=0.483738\textwidth]{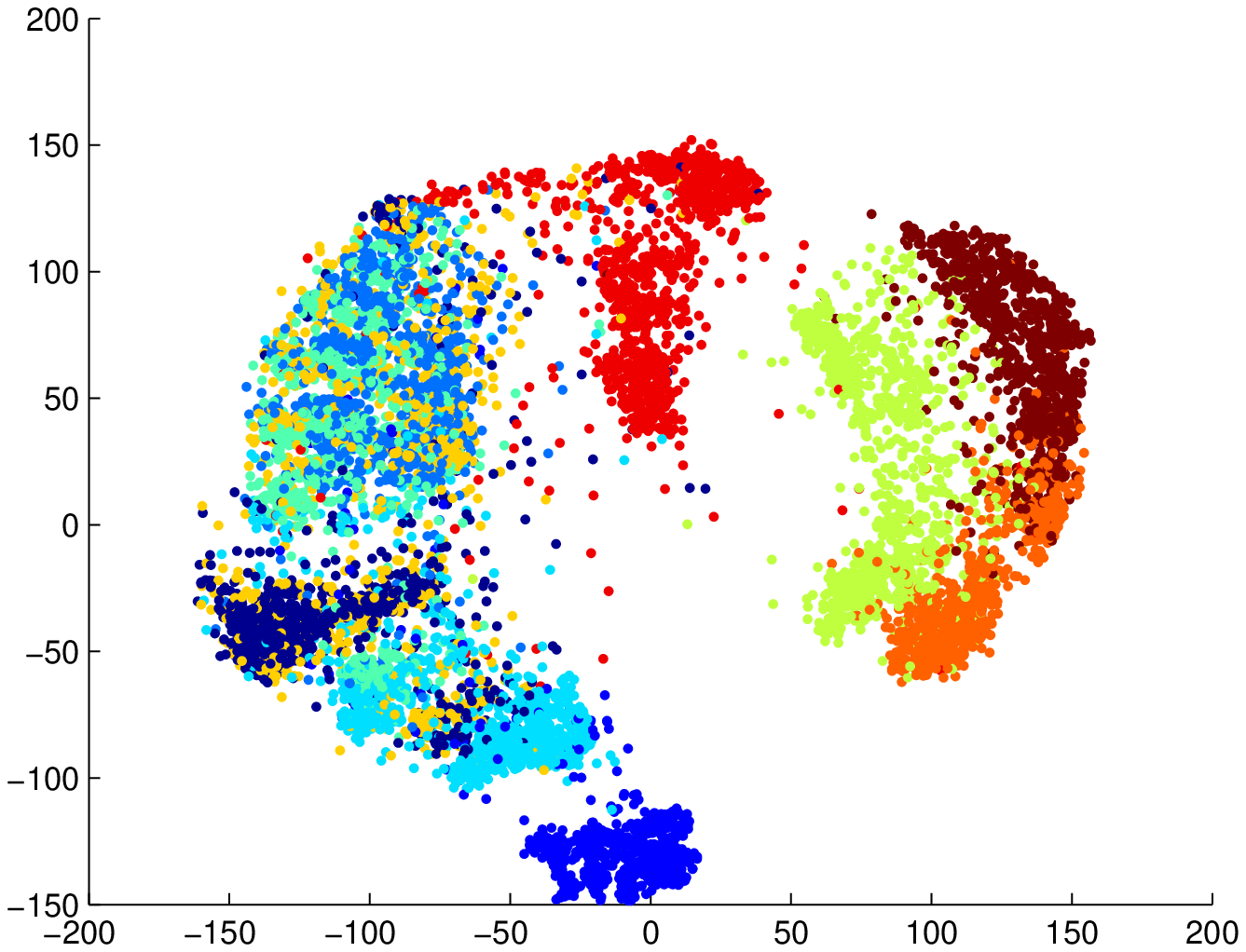}
    }
  \centering
  \caption{Fashion embedding figures for pt-SNE and hot-SEE}
  \label{fig:fashionEmbedding}
  \end{figure}
Also, in Figures~\ref{fig:fashionEmbedding}, we depicted our visual 2D embedding results on the Fashion data set. These figures further confirmed the better clustering quality generated by the hot-SEE method, compared to that of the pt-SNE strategy.

\subsection{Noise Contrastive Estimation}
In this section, we evaluated the performance of the noise contrastive estimation (NCE) strategy applied to our method hot-SEE with perplexity 3 and 2000 exemplars. We set $z_e = z_n = 100$ and $K_e = 18$. Table~\ref{tab:nce} show the error rates (\%) obtained by 1NN on the two-dimensional representations produced by hot-SEE with or without NCS, respectively, on the MNIST and Fashion datasets.

\begin{table}[H]
  \centering
 \begin{tabular}{|lc|lc|lc|lc|}\hline
\multicolumn{4}{|c}{MNIST}&\multicolumn{4}{|c|}{Fashion}\\ \hline
\multicolumn{2}{|c|}{standard}&\multicolumn{2}{c|}{w/ NCE} &\multicolumn{2}{c|}{standard}&\multicolumn{2}{c|}{w/ NCE}  \\
\hline 
\multicolumn{2}{|c|}{9.30}&\multicolumn{2}{c|}{9.69} &\multicolumn{2}{c|}{28.18}&\multicolumn{2}{c|}{28.19}\\
\hline
\end{tabular}
\caption{Error rates (\%) obtained by 1NN on the two-dimensional representations produced by hot-SEE (perplexity = 3 and 2000 exemplars) with or without further computational complexity reduction based on Noise Contrastive Estimation (NCE), respectively, on the MNIST and Fashion datasets.} \label{tab:nce}
\end{table}

Results in Table~\ref{tab:nce} suggest that the NCE was able to further reduce the 
  computational and memory complexity of our method without sacrificing the predictive performance. As shown in the table, the  accuracy difference of the hot-SEE method with and without NCE was less than 0.4\% for both the MNIST and Fashion data sets. 

\begin{table*}[h]
  \centering

	\scalebox{0.95}{    
 \begin{tabular}{|lc|lc|lc|lc|lc|lc|}\hline
\multicolumn{4}{|c|}{COIL100}&\multicolumn{4}{c}{MNIST}&\multicolumn{4}{|c|}{Fashion}\\ \hline
\multicolumn{2}{|c|}{careful seeding}&\multicolumn{2}{c|}{random seeding} &\multicolumn{2}{c|}{careful seeding}&\multicolumn{2}{c|}{random seeding} &\multicolumn{2}{c|}{careful seeding}&\multicolumn{2}{c|}{random seeding} \\
\hline 
\multicolumn{2}{|c|}{58.67}&\multicolumn{2}{c|}{58.44} &\multicolumn{2}{c|}{9.30}&\multicolumn{2}{c|}{9.19} &\multicolumn{2}{c|}{28.18}&\multicolumn{2}{c|}{28.53} \\
\hline
\end{tabular}}
  \caption{Error rates (\%) obtained by 1NN on the 2-dimensional representations produced by hot-SEE (perplexity = 3) with careful seeding or random seeding on the COIL100 (with 600 exemplars), MNIST (with 2000 exemplars), and Fashion (with 2000 exemplars) datasets.}
  \label{tab:seed}
\end{table*}
  
\subsection{Careful Exemplar Seeding vs. Random Initialization}
We also further evaluate the performance of our methods in terms of different exemplar initializations used. We compared the performance of using careful seeding based on scalable K-means++ and randomly initialized exemplars. We presented the results in Table~\ref{tab:seed}. From Table~\ref{tab:seed}, one can observe that  our methods were insensitive to the exemplar seeding approach used. That is, very similar predictive performances (less than 0.4\%) were obtained by our methods on all the three testing data sets, i.e., COIL100, MNIST, and Fashion. 

\subsection{Comparing Evaluation Metrics of kNN ($k \geq 1$) and Quality Score}
We believe that the evaluation metric based on 1NN test error rate used in the previous experimental sections is more appropriate than kNN test error rate with $k > 1$. The reason is that the 1NN performance exactly shows how accurately our exemplar-based embedding methods catpure very local neighborhood information, which is more challenging for our proposed methods. Because exemplars are computed globally, it is much easier for dt-see and hot-see to achieve better performance based on kNN with $k > 1$. On the MNIST dataset, we show the best training and test error rates of kNN with $k \geq 1$ using the two-dimensional embedding generated by different methods in Talbe~\ref{tab:knnerr}, which consistently shows that dt-see and hot-see significantly outperforms pt-SNE and supports our claims above.
\begin{table*}[h]
  \centering

	\scalebox{0.95}{    
 \begin{tabular}{|l|c|c|c|c|c|c|c|c|c|c|}
\hline
&\multicolumn{10}{c|}{The Number of Nearest Neighbors k in kNN}\\
\hline
Method&1&2&3&4&5&6&7&8&9&10\\
\hline
pt-sne\_tr&12.49&12.49&9.26&8.84&8.45&8.30&8.18&8.12&8.08&8.08\\
pt-sne\_te&12.55&12.55&9.79&9.48&9.12&8.95&8.83&8.72&8.72&8.69\\
\hline
hot-see\_tr&8.87&8.87&6.31&6.05&5.83&5.68&5.64&5.63&5.60&5.58\\
hot-see\_te&9.19&9.19&7.21&6.76&6.61&6.42&6.41&6.41&6.42&6.36\\
\hline
dt-see\_tr&\textbf{7.19}&\textbf{7.19}&\textbf{5.09}&\textbf{4.90}&\textbf{4.72}&\textbf{4.67}&\textbf{4.62}&\textbf{4.62}&\textbf{4.56}&\textbf{4.56}\\
dt-see\_te&\textbf{8.80}&\textbf{8.80}&\textbf{6.69}&\textbf{6.45}&\textbf{6.25}&\textbf{6.17}&\textbf{6.02}&\textbf{6.02}&\textbf{5.94}&\textbf{5.96}\\
\hline
\end{tabular}}
  \caption{The training error rates (\_tr) and test error rates (\_te) of kNN with different k's using the two-dimensional embedding generated by different methods on MNIST.}
  \label{tab:knnerr}
\end{table*}

Another evaluation metric based on Quality Score was used by a recent method called kernel t-SNE (kt-SNE) \cite{ktsne}. The Quality Score metric computes the k (neighborhood size) nearest neighbors of each data point, respectively, in the high-dimensional space and in the low-dimensional space, and the metric calculates the preserved percentage of the high-dimensional neighborhood in the low-dimensional neighborhood averaged over all test data points as the Quality Score, with respect to different neighborhood size k. In Table~\ref{tab:qualityscores}, we compute the quality scores of different methods on the MNIST test data for preserving their neighborhood on the training data with neighborhood size ranging from 1 to 100. These results also show that hot-see and dt-see consistently outperform pt-SNE. 

We find that Kernel t-SNE is also capable of embedding out-of-sample data. To have a similar experiment setting on MNIST as that used in kernel t-SNE, we randomly choose 2000 data points as held-out test set from the original test set (size=10000) to get 10 different test sets with size 2000, the test error rates of our methods compared to kernel t-SNE are, kernel t-SNE: $14.2\%$, fisher kernel t-SNE: $13.7\%$, hot-see: $9.11\% \pm 0.43\%$, dt-see: $8.74\% \pm 0.37\%$. Our methods hot-see and dt-see significantly outperform (fisher) kernel t-SNE.

\begin{table*}[h]
  \centering

	\scalebox{0.95}{    
 \begin{tabular}{|l|c|c|c|c|c|c|c|c|c|c|c|c|}
\hline
&\multicolumn{11}{c|}{Neighborhood Size}\\
\hline
Method&1&10&20&30&40&50&60&70&80&90&100\\
\hline
pt-sne&0.55&4.01&6.68&8.76&10.56&12.17&13.62&14.93&16.06&17.19&18.23\\
hot-see&1.12&5.25&8.22&10.53&12.48&14.19&15.69&17.04&18.27&19.41&20.44\\
dt-see&\textbf{1.14}&\textbf{6.74}&\textbf{10.68}&\textbf{13.52}&\textbf{15.78}&\textbf{17.56}&\textbf{19.03}&\textbf{20.22}&\textbf{21.31}&\textbf{22.27}&\textbf{23.17}\\
\hline
\end{tabular}}
  \caption{Quality scores (\%, the higher the better) for different embedding methods computed on the test set against the training set on MNIST.}
  \label{tab:qualityscores}
\end{table*}

\section{Conclusion and Future Work}
\label{sec:discussion}
In this paper, we present unsupervised parametric t-distributed stochastic exemplar-centered data embedding and visualization approaches, leveraging a deep neural network or a shallow neural network with high-order feature interactions. 
Owing to the benefit of a small number of precomputed high-dimensional exemplars, our approaches avoid pairwise training data comparisons and have signicantly reduced computational cost. In addition, the high-dimensional exemplars reflect local data density distributions and global clustering patterns. With these nice properties, the resulting embedding approaches solved the important problem of embedding performance being sensitive to hyper-parameters such as batch sizes and perplexities, which have haunted other neighbor embedding methods for a long time. Experimental results on several benchmark datasets demonstrate that our proposed methods significantly outperform state-of-the-art unsupervised deep parametric embedding method pt-SNE in terms of robustness, visual effects, and quantitative evaluations.

In the future, we plan to incorporate recent neighbor-embedding speedup developments based on efficient N-body force approximations into our exemplar-centered embedding framework. 

\bibliographystyle{splncs04}
\bibliography{ref}
\end{document}